\documentclass{article}

\usepackage{arxiv}
\usepackage{amsmath}
\usepackage[utf8]{inputenc} 
\usepackage[T1]{fontenc}    
\usepackage{hyperref}       
\usepackage{url}            
\usepackage{booktabs}       
\usepackage{amsfonts}       
\usepackage{nicefrac}       
\usepackage{microtype}      
\usepackage{lipsum}
\usepackage{graphicx}
\graphicspath{ {./images/} }

\title{Validation of Human Pose Estimation and Human Mesh Recovery for Extracting Clinically Relevant Motion Data from Videos}

\author{
  Kai Armstrong \\
  School of Computer Science \\
  University of Lincoln \\
  \texttt{karmstrong@lincoln.ac.uk} \\
  \And
  Alexander Rodrigues \\
  School of Psychology, Sport Science, and Wellbeing \\
  University of Lincoln \\
  \texttt{arodrigues@lincoln.ac.uk} \\
  \And
  Alexander P. Willmott \\
  School of Psychology, Sport Science, and Wellbeing \\
  University of Lincoln \\
  \texttt{swillmott@lincoln.ac.uk} \\
  \And
  Lei Zhang \\
  School of Computer Science \\
  University of Lincoln \\
  \texttt{lzhang@lincoln.ac.uk} \\
  \And
  Xujiong Ye \\
  School of Computer Science \\
  University of Lincoln \\
  \texttt{xye@lincoln.ac.uk} \\
}

\begin{document}
\maketitle
\begin{abstract}
This work aims to discuss the current landscape of kinematic analysis tools, ranging from the state-of-the-art in sports biomechanics such as inertial measurement units (IMUs) and retroreflective marker-based optical motion capture (MoCap) to more novel approaches from the field of computing such as human pose estimation and human mesh recovery. Primarily, this comparative analysis aims to validate the use of marker-less MoCap techniques in a clinical setting by showing that these marker-less techniques are within a reasonable range for kinematics analysis compared to the more cumbersome and less portable state-of-the-art tools. Not only does marker-less motion capture using human pose estimation produce results in-line with the results of both the IMU and MoCap kinematics but also benefits from a reduced set-up time and reduced practical knowledge and expertise to set up. Overall, while there is still room for improvement when it comes to the quality of the data produced, we believe that this compromise is within the room of error that these low-speed actions that are used in small clinical tests.  
\end{abstract}


\section{Introduction}
The field of motion capture technologies is a rapidly advancing area with multiple directions, ranging from capturing human motion using infra-red (IR) cameras with retro-reflective markers \cite{guerra2005optical}, to wearable Inertial Measurement Units (IMUs) \cite{prayudi2012design}. More recently, there have been advances in marker-less motion capture technologies, ranging from portable solutions that can run on a single smartphone to a more complex arrangement of cameras to produce a more accurate result \cite{desmarais2021review, wren2023comparison}. The purpose of the study is to identify whether the use of marker-less techniques can be used over the current state-of-the-art and to identify the areas to which this technology is most suited. 

Optical motion capture (MoCap) is considered by most to be the gold standard for motion analysis systems, the customizable approach allows any number of kinematics and biomechanics to be measured from a subject in conjunction with both high precision and high frequency data collection \cite{fern2012biomechanical}. However, cheaper options are beginning to become available, such as the DeMoCap system, which uses a reduced set of depth cameras with standard retroreflective marker sets \cite{chatzitofis2021democap}. The gold standard systems range from \$5000-250,000 depending on the specification and requirements, these can also have additional associated and indirect costs such as a dedicated room that could not be used for other uses and having a dedicated technician with training and experience in MoCap \cite{carse2013affordable}. 

 An alternative to optical motion capture is the inertial measurement unit (IMU), a combination of accelerometer, gyroscope and magnetometer, these sensors are used to track the change in motion and rotation of different limbs, thus allowing calculation of joint kinematics and biomechanics \cite{hamdi2014lower}.  Although these sensors are significantly cheaper and allow data collection outside of a laboratory environment than a full optical MoCap suite, to collect the same level of information, more sensors are required. However, this modular approach makes these devices a more affordable option for clinical environments, for example a clinic would often not require both upper and lower limb kinematics of both sides of the body simultaneously \cite{gu2023imu}. 

Marker-less MoCap, as opposed to the previously discussed techniques, relies on a learned approach to understanding the anatomy and kinematics of humans. These models are trained on tens to hundreds of thousands of images and videos, then a prediction is made to the unknown image and estimate the location of joints within the body in respect to the image co-ordinates \cite{bazarevsky2020blazepose, pavlakos2019expressive}. Some models can be run with real-time pose estimation on low-cost devices such as smartphones \cite{kulkarni2023poseanalyser}, however, other more advanced methods require extensive computing power a more expensive graphics processing unit (GPU) \cite{kocabas2020vibe, li2021hybrik}. 

These vision-based techniques do come with their own set of limitations. For example, depending on the placement of the cameras and the way in which people can perform certain actions, some points on the body can be self-occluded which can lead to an incorrect landmark position estimation or a "jitter" between frames \cite{mundermann2006evolution}. Additionally, similarly to the marker-based MoCap, it has been found that the choice of clothes can affect the predicted pose \cite{matsumoto2020human} which will require additional experiments to test this observation further. Additionally, it has been found that the accuracy of the pose prediction can be affected by the lighting of the environment which has also been discussed by Ye \textit{et al.} \cite{cheng2015capturing, ye2022effects}. 

Out of the described techniques, both the IMU and marker-less MoCap have a viable use in-the-wild and in a clinical setting due to their inherent speed and portability \cite{zhou2020we, stenum2021applications}. This speed and portability is useful to deal with the fundamental problems faced in the healthcare industry, the access to healthcare whether this is the portability to bring the tools to more rural areas or the speed to see more patients faster \cite{islam2019portable, petersen2021systematic}. However, between the two of these techniques the IMUs are preferred as this is not constrained by the camera position and distance from the subject. 

Overall, there are many considerations that must be taken into account when using techniques and hardware from sports biomechanics in a clinical setting. As very few methods have been optimized for small and less than ideal data collection environments, we must look towards the rapidly advancing fields of marker-less human motion analysis methods. However, there are a limited number of studies that have directly assessed the capabilities of these methods for their accuracy when compared to the gold standards in sports biomechanics while also discussing the potential for clinical applications. Thus, this study aims to validate the use of marker-less technologies for use in a clinical decision-making pipeline by focusing on precision in a previously validated clinically relevant set of actions \cite{armstrong2024marker}.

\section{Materials and Methods}

Data collection was arranged with the Human Performance Centre (HPC), University of Lincoln School of Psychology, Sport Science and Wellbeing, who provided the hardware and the location for collecting all data. This hardware consisted of two Sony DSC-RX0 II 4k high frame rate (HFR) cameras, six Vicon Blue Trident IMU devices, and a total of 12 Motion Analysis Corporation Raptor cameras of varying models placed around the room. The dataset also contains ground reaction force data from two Kistler force plates; this data was not used for the subsequent data analysis as this is beyond the scope of the study but is available for future studies. 

\subsection{Data Collection}

In total there were 10 participants in this trial, recruited from within the research institution. Participant recruitment consisted of "healthy" individuals, defined as those without diagnosed knee conditions, all participants were above the age of 18 and consented to have their data collected in accordance with the University of Lincoln’s Ethics Approval. Patient demographic statistics are shown in Table \ref{tab:patientdems}. 

\begin{table}
  \centering
  \begin{tabular}{lccc}
    \toprule
    Statistic & Mean & Standard Deviation & Range \\
    \midrule
    Height (m) & 177.2 & 4.4 & 10.7 \\
    Mass (kg)  & 69.8  & 7.1 & 20.9 \\
    BMI        & 21.1  & 4.4 & 6.3  \\
    \bottomrule
  \end{tabular}
  \caption{Distribution of participants for the study showing the mean, standard deviation, and range for relevant statistics such as height, weight, and BMI.}
  \label{tab:patientdems}
\end{table}

Participants were asked to perform a series of actions in two test conditions, "normal” clothing consisting of loose-fitting clothing such as jeans and sweatshirts and "MoCap" - clothing consisting of tight-fitting and dark clothing, both test conditions were performed on the same day with a break of 15 minutes between to rest and reset the marker and IMU placement. 

The ordered list of recorded actions is as follows: 
\begin{itemize}
    \item Static, n=1;
    \item Squat, n=5;
    \item Lunge (left and right), n=3;
    \item Single leg balance (left and right), n=1;
    \item Squat to box, n=5;
    \item Sit to stand, n=3.
\end{itemize}

A total of 24 retroreflective markers (12.5 mm diameter) were placed on specific landmarks ahead of data collection. These were more accurately placed on skin or tight-fitting clothing in the “MoCap” condition, compared to placing over the top of participant’s clothes in the “Normal Clothes” (NC) condition. For consistency, participants were asked to remain standing during marker application in the NC condition. The markers were placed on the left and right ASIS, PSIS, greater trochanter, lateral and medial femoral epicondyles, lateral and medial malleoli, first and fifth metatarsal heads as well as on either thigh and shank (as offset markers). Markers were tracked using a 12-camera cortex system, consisting of x Raptor-4 cameras and x Raptor-E cameras, recording at a frequency of 150 Hz. Following laboratory changes, an additional Raptor-X camera, also recording at 150Hz, was used for Participant 10

The six Inertial Measurement Units (Vicon Blue Trident) were placed directly onto participant’s skin and shoes, on the lateral side of each segment (Left and Right Foot, Shank and Thigh) operating at 300 Hz in each trial. For the NC condition, the participant’s clothes were placed over the top of these sensors. This data was collected straight to the sensors (firmware version 10.0.1) using an iPad Pro and the Vicon Capture.U application (version 1.4.0.9852). Data was then transferred to the computer following each data collection 
session.  

\begin{figure}
    \centering
    \includegraphics[width=1\linewidth]{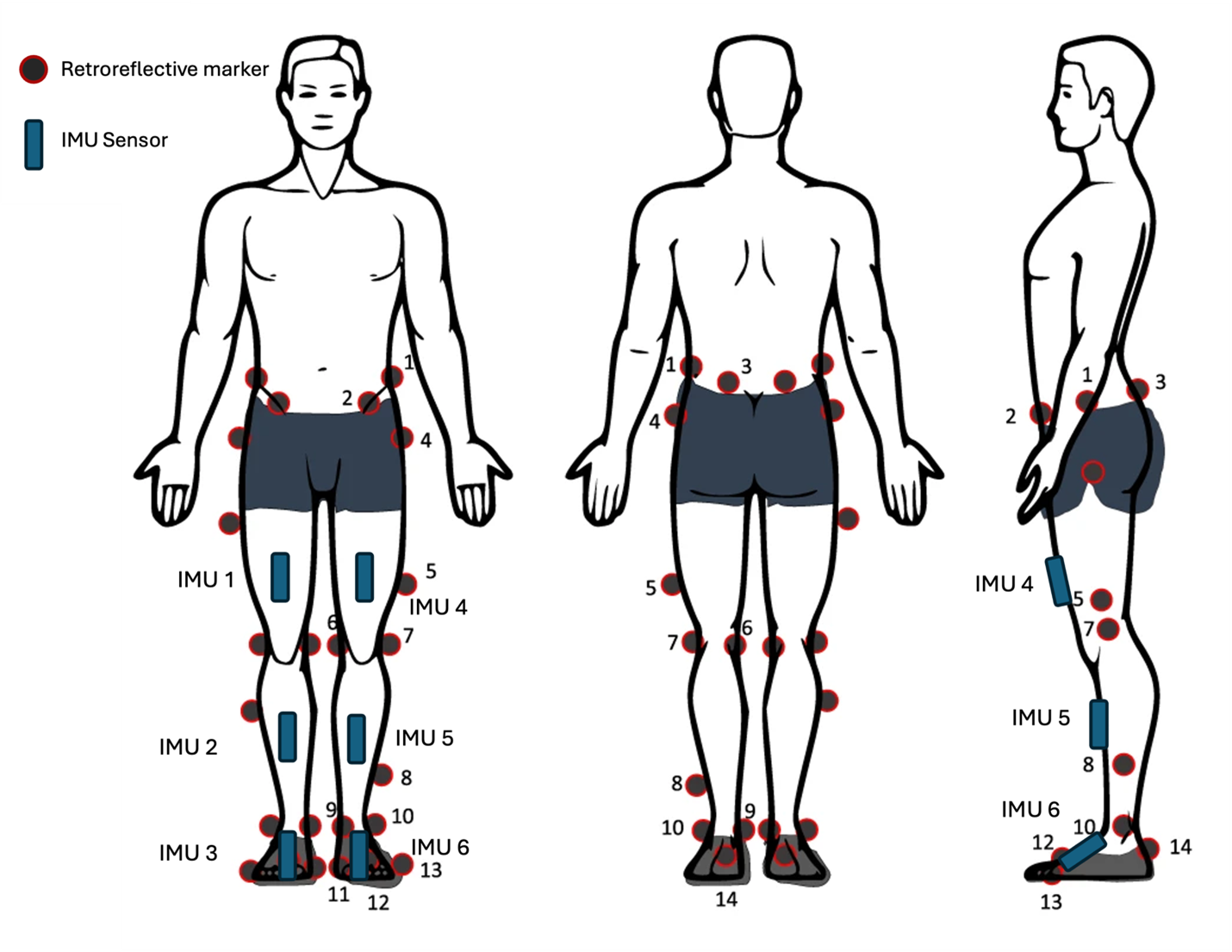}
    \caption{Diagram showing the placement of the retroreflective MoCap markers and IMU sensors in the sagittal, coronal and posterior views; allowing for lower limb kinematics, adapted from Reznick et al. 2021 \cite{reznick2021lower}.}
    \label{fig:marker-placement}
\end{figure}

These retroreflective markers and IMU sensor locations can be seen in Figure \ref{fig:marker-placement}, these positions were adapted from Reznick et al. 2021 \cite{reznick2021lower}. 

The two tripod mounted Sony Cameras (recording at 30 Hz), MoCap System and IMU’s were all used to record each trial.  All cameras were connected to allow for near-synchronous starts of recording. Since the IMUs couldn’t be directly synced to the MoCap system, participants were instructed to only begin actions at very particular times, giving the system operators time to ensure the IMUs were recording the same 
actions as the other systems.

\subsection{Motion Capture Processing}

Raw MoCap data was initially processed using Cortex-64 (Version 9.0.0.44) to assign each marker to their respective body segment locations, joint marker coordinate data were then smoothed using a Savitsky Golay filter in the Matlab Signal Processing Toolbox; with a window length of 15 and a polynomial order of 3.

Joint angles were then calculated using the vector cross product method, as detailed in Equations \ref{eq:cross}, \ref{eq:dot} and \ref{eq:theta}. This method calculates the knee angle, for example, $\theta$ by first extracting the thigh and shank vectors, $T$ and $S$, based on the relevant joint centres. We then compute the angle using the cross product $\mid C \mid$ and the dot product $T\cdot S$ to calculate the angle in the correct quadrant.

\begin{equation}
    \vec{T} = \text{Hip} - \text{Knee} \quad \vec{S} = \text{Knee} - \text{Ankle}
    \label{eq:cross}
\end{equation}

\begin{equation}
    \vec{C} = \vec{T} \times \vec{S} = 
    \begin{vmatrix} 
    \hat{i} & \hat{j} & \hat{k} \\
    T_x & T_y & T_z \\
    S_x & S_y & S_z
    \end{vmatrix}
    \label{eq:dot}
\end{equation}

\begin{equation}
    \theta = \arctan2 \left( |\vec{C}|, \vec{T} \cdot \vec{S} \right)
    \label{eq:theta}
\end{equation}

\subsection{Inertial Measurement Unit Processing}

IMU sensor processing was processed using the open-source GitHub repository "Biomech Zoo"(\url{https://github.com/PhilD001/biomechZoo}) \cite{dixon2017biomechzoo}.
The provided joint angle calculation first takes a pair of any given segments, the knee angle for example we take the right thigh and shank sensors, the function will determine each segment's orientation as a quaternion. The quaternion of the knee joint, $Q_{\text{knee}}$, is calculated using the relative orientation between the thigh and shank quaternions $Q_{\text{thigh}}$ and $Q_{\text{shank}}$ as denoted in equation \ref{eq:to_quat}.
\begin{equation}
Q_{\text{knee}} = \text{conj}(Q_{\text{thigh}}) \cdot Q_{\text{shank}}
\label{eq:to_quat}
\end{equation}
Where conj denotes the quaternion conjugate.
We then convert the quaternion into Euler angles and convert the knee angle into radians to allow for consistency in the measurements, this is shown in Equation \ref{eq:to_euler}.

\begin{equation}
\theta = -\text{rad2deg}(Q_{\text{knee},1})
\label{eq:to_euler}
\end{equation}

\subsection{RGB Video Processing}

Videos were processed using two different vision-based motion extraction methods, human pose estimation using Mediapipe and human mesh reconstruction using the Video Inference for Human Body Pose and Shape Estimation (VIBE) model. Firstly, for the human pose estimation we used Mediapipe heavy pose landmarker model, creating a 33-joint skeleton system with positions in three-dimensional Cartesian space whereby the co-ordinates represent the real distance in meters relative to the centre of the two hip joints \cite{bazarevsky2020blazepose}. This skeleton system was created for each frame in the video creating a 3D skeleton sequence.

The videos were also processed into a Skinned Multi-Person Linear Model (SMPL) mesh \cite{loper2023smpl} using the VIBE model \cite{kocabas2020vibe}. The SMPL model represents the surface mesh with a human skeleton system inside, this skeleton system consists of 49 3D joints in root relative space similarly to mediapipe.
After processing the videos into pose sequences, the joint co-ordinates were also smoothed using a low-pass Savitzky-Golay filter was applied using a window size of 5 and a polynomial order of 1. These smoothed pose sequences were then used to calculate the joint angle kinematics, all joint angles were calculated using the cosine equation \ref{eq:cosine}. Given the primary goal of determining the accuracy of marker-less motion capture systems for the knee, the knee, hip and ankle were used as k, h, and a respectively.

\begin{equation}
\theta = \cos^{-1} \left( \frac{|m|^2 + |n|^2 - |p|^2}{2|m|\cdot|n|} \right)
\label{eq:cosine}
\end{equation}

For vectors: $\vec{m} = \vec{h} \rightarrow \vec{k}$, $\vec{n} = \vec{a} \rightarrow \vec{k}$ and $\vec{p} = \vec{h} \rightarrow \vec{a}$

where $\vec{k}$, $\vec{h}$, and $\vec{a}$, are the positions of the joint centers of any three given joints, for example knee, hip, and ankle respectively can be used to calculate the knee angle and $|x|$ denotes the Euclidean distance between two points.

\subsection{Data Analysis}

The analysis of the generated motion data has two primary avenues for analysis, on the individual level and the population level, each with different methodologies and objectives.

\subsubsection{Individual-Level Analysis}

To compare joint angle kinematics across all data modalities, we employed a normalization and alignment process. First, data from all modalities were resampled to ensure a consistent sampling frequency of 10Hz across the x-axis. This normalization enabled direct comparison between modalities. Given the asynchronous nature of data collection technologies and varying user reaction speeds, the raw kinematic data lacked a consistent start and end time. To address this, we manually shifted the time-series curves along the x-axis. This alignment was performed by identifying and matching key action points, such as the initiation or peak of a movement. This qualitative approach ensured that temporal discrepancies were minimized, allowing us to assess the overall agreement between modalities. The resulting aligned data provided a foundation for identifying potential discrepancies or biases across modalities.

\subsubsection{Population-Level Analysis}

The population-level analysis involved a multi-step pipeline designed to evaluate and quantify the agreement between data modalities in measuring joint angle kinematics. This pipeline first extracted key kinematic metrics of the left and right knee flexion. These metrics included the minimum knee angle, maximum knee angle, and range of motion (ROM). For each participant and data modality, these values were computed to establish a baseline comparison.

To assess the agreement and systematic bias between data modalities, Bland-Altman plots were generated for the minimum knee angle, maximum knee angle, and ROM metrics. These plots provided a visual representation of the agreement and highlighted any systematic differences. The mean difference and limits of agreement (mean ± 1.96 standard deviations) were calculated to quantify bias and variability.

Errors associated with kinematic metrics were quantified using the mean absolute error (MAE) and mean signed error (MSE) as shown in Equations \ref{eq:mae} and \ref{eq:mse}.

\begin{equation}
\text{MAE} = \frac{1}{n} \sum_{i=1}^{n} |y_i - \hat{y}_i|
\label{eq:mae}
\end{equation}

\begin{equation}
\text{MSE} = \frac{1}{n} \sum_{i=1}^{n} (y_i - \hat{y}_i)
\label{eq:mse}
\end{equation}
Where:
\begin{itemize}
\item $n$ is the number of observations
\item $y_i$ is the actual value for the $i$-th observation
\item $\hat{y}_i$ is the predicted value for the $i$-th observation
\item $|y_i - \hat{y}_i|$ is the absolute value of the error
\item $(y_i - \hat{y}_i)$ is the signed error (can be positive or negative)
\end{itemize}

Here, represents the metric value for a given modality, and represents the reference value. These metrics were computed for the sit-to-stand and squat-to-box actions, chosen due to their consistent and well-defined movement boundaries. Unlike free squats, which vary across individuals, these actions provided standardized reference points, such as the consistent height of a chair.

To ensure the modalities were statistically comparable, Pearson correlation coefficients were calculated for the kinematic metrics, providing a measure of linear relationship strength between modalities. Additionally, paired t-tests were performed to detect any statistically significant differences in measurements between modalities. These tests were conducted for both clothing conditions and across multiple camera angles for the marker-less MoCap method, ensuring comprehensive evaluation of the dataset.

For the marker-less MoCap method, specific attention was given to the influence of clothing and camera angle. Each data collection session involved two clothing sets—tight-fitting and loose-fitting—to determine how clothing affected the accuracy of the captured kinematics. Furthermore, the motion data from two camera angles (frontal and sagittal) were evaluated to understand their impact on measurement reliability.
The evaluation of these factors was integrated into both individual-level and population-level analyses to account for potential variability introduced by these parameters.

\section{Results}

\subsection{Individual Analyses}

\begin{figure}
    \centering
    \includegraphics[width=1\linewidth]{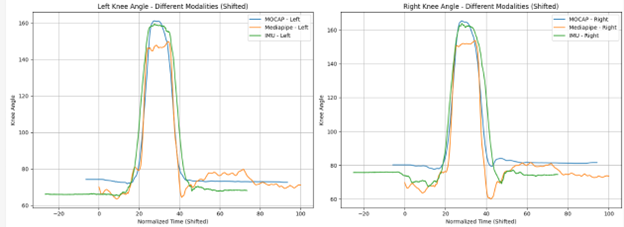}
    \caption{Resampled and shifted plot of the left and right knee flexion curves comparing the MoCap, IMU, and Mediapipe-based joint angles for the "MoCap friendly”clothing in a sit-to-stand action}
    \label{fig:p06-sts}
\end{figure}

\begin{figure}
    \centering
    \includegraphics[width=1\linewidth]{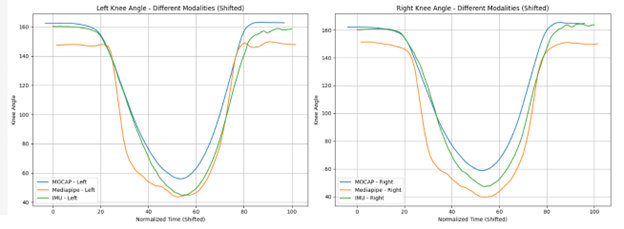}
    \caption{Resampled and shifted plot of the left and right knee flexion curves comparing the MoCap, IMU, and Mediapipe-based joint angles for the "MoCap friendly”clothing in a squat action}
    \label{fig:p06-sqt}
\end{figure}

\begin{figure}
    \centering
    \includegraphics[width=1\linewidth]{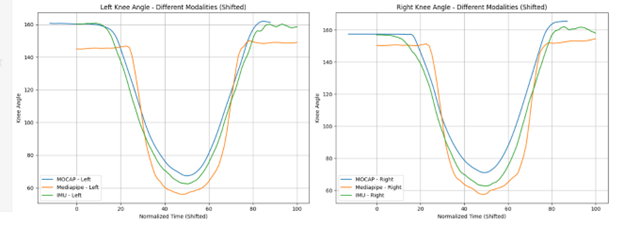}
    \caption{Resampled and shifted plot of the left and right knee flexion curves comparing the MoCap, IMU, and Mediapipe-based joint angles for the "MoCap friendly”clothing in a squat to box action}
    \label{fig:p06-sqb}
\end{figure}

The results of the individual comparison of modalities in Figures \ref{fig:p06-sts}, \ref{fig:p06-sqt} and \ref{fig:p06-sqb} are the sit-to-stand, squat to box and squat actions from participant 6; these results are representative of each participant in the population. From Figure \ref{fig:p06-sts} we see that during the sit-to-stand mediapipe and the IMU start and end at a similar knee angle but mediapipe’s peak knee angle is lower than that of both the IMU and MoCap. On the other hand, in Figure \ref{fig:p06-sqt} we see that the mediapipe and the IMU share the same lowest knee angle whereas the start and the end of the action are lower than both the IMU and MoCap knee angle. Similarly in \ref{fig:p06-sqb} mediapipe starts and ends roughly 15° below the IMU and MoCap, however, in this action none share a similar lowest knee angle value. 

\subsection{Generalised Overview}

\begin{figure}
    \centering
    \includegraphics[width=1\linewidth]{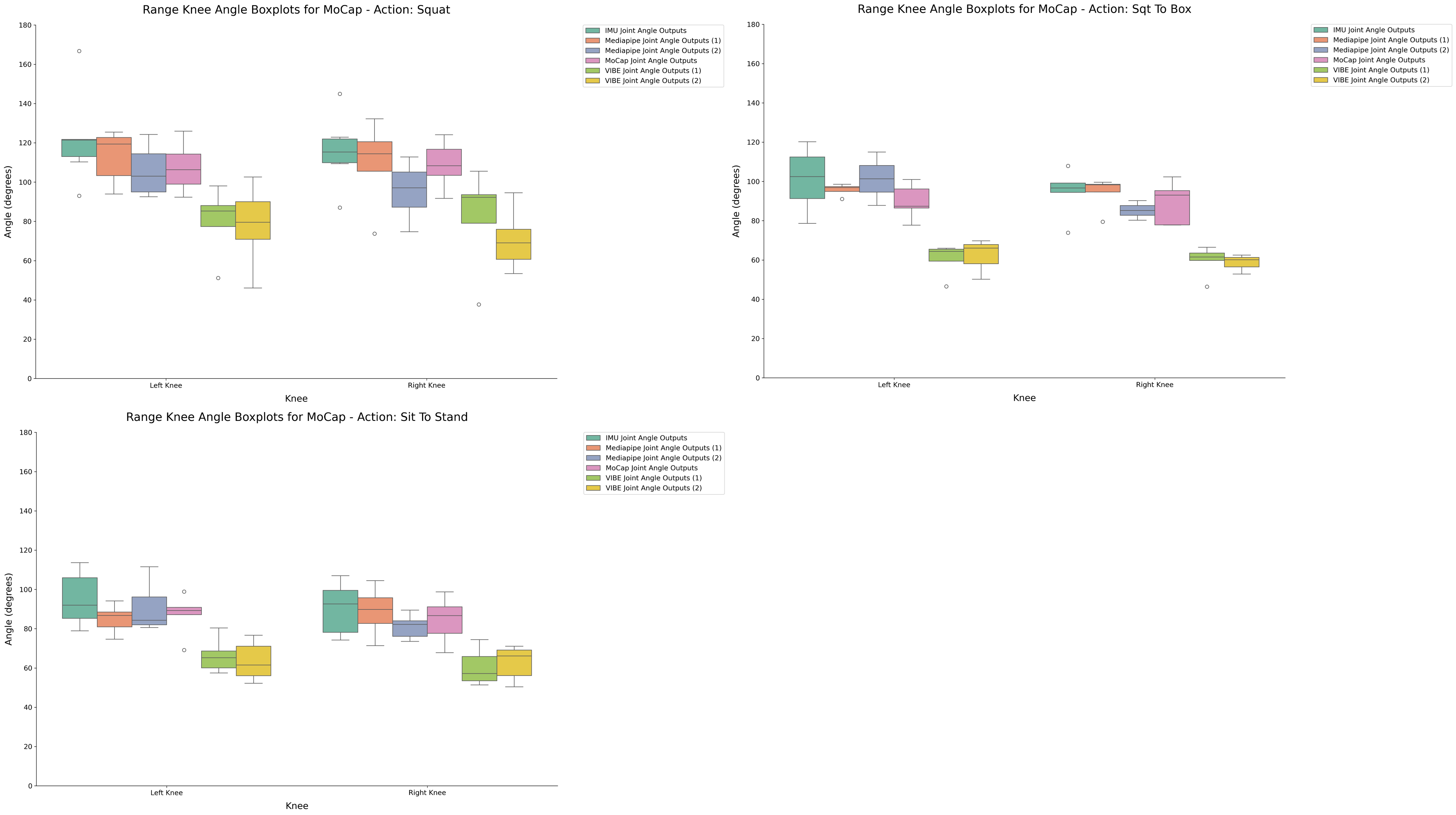}
    \caption{Boxplots showing the range of motion for the left and right knees while wearing MoCap friendly clothing measured during the sit-to-stand, squat to box, and squat actions.}
    \label{fig:boxplot_mocap}
\end{figure}

Figures \ref{fig:boxplot_mocap} and y2 show the range of motion (RoM) of both the left and right knee flexion, where \ref{fig:boxplot_mocap} represents the kinematics in “MoCap friendly” clothing and y2 shows the kinematics measured while wearing “Normal” clothing. These figures indicate that for all actions and clothing conditions, the MediaPipe kinematics, MoCap kinematics, and IMU-based kinematics are in line with one another for the RoM. Meanwhile, the VIBE-based kinematics deviate from the other modalities for the RoM. Additionally, the results in figure y2 show a wider spread in the data compared to figure \ref{fig:boxplot_mocap}.

\begin{figure}
    \centering
    \includegraphics[width=1\linewidth]{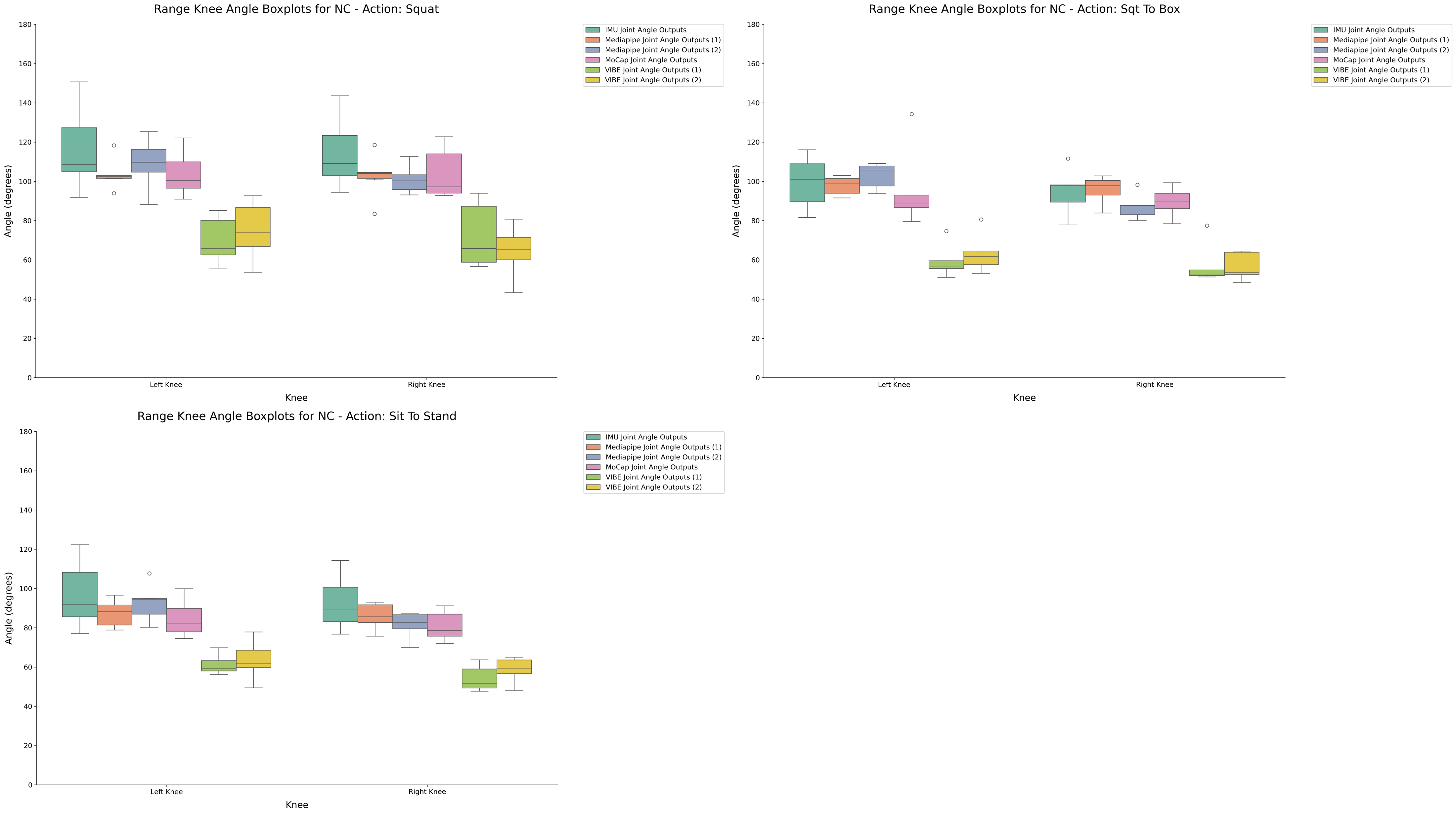}
    \caption{Boxplots showing the range of motion for the left and right knees while wearing normal clothing measured during the sit-to-stand, squat to box, and squat actions.}
    \label{fig:enter-label}
\end{figure}

\begin{figure}
    \centering
    \includegraphics[width=1\linewidth]{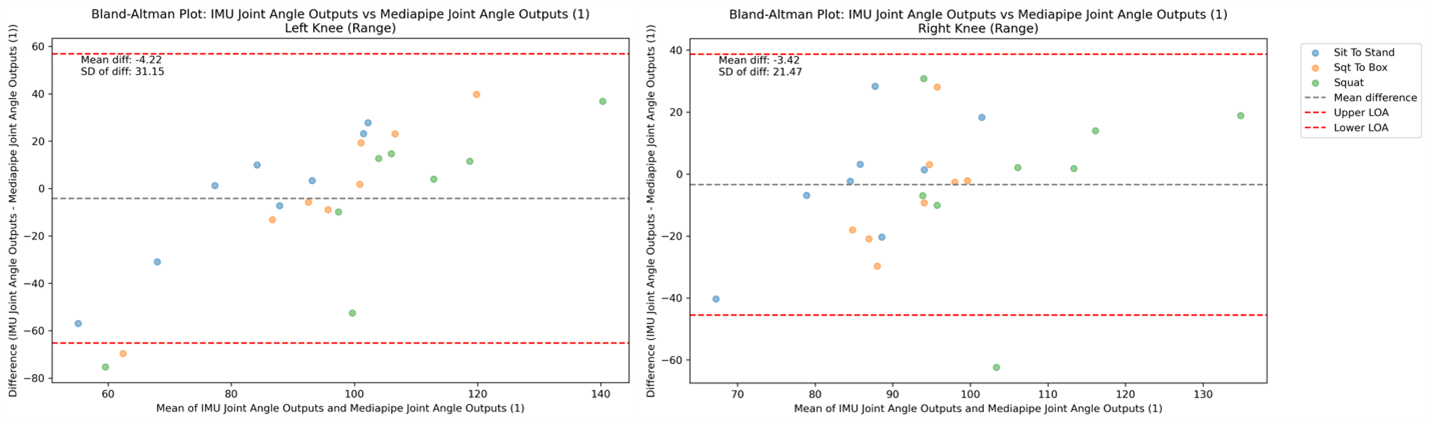}
    \caption{Bland-Altman plot illustrating the agreement between Mediapipe and IMU. The mean difference (bias) is shown as a solid line, while the dashed lines represent the limits of agreement (mean difference ± 1.96 standard deviations), the points represent the range of motion of both the left knee flexion (left) and right knee flexion (right) of different actions for each participant.}
    \label{fig:ba_mediapipe}
\end{figure}

\begin{figure}
    \centering
    \includegraphics[width=1\linewidth]{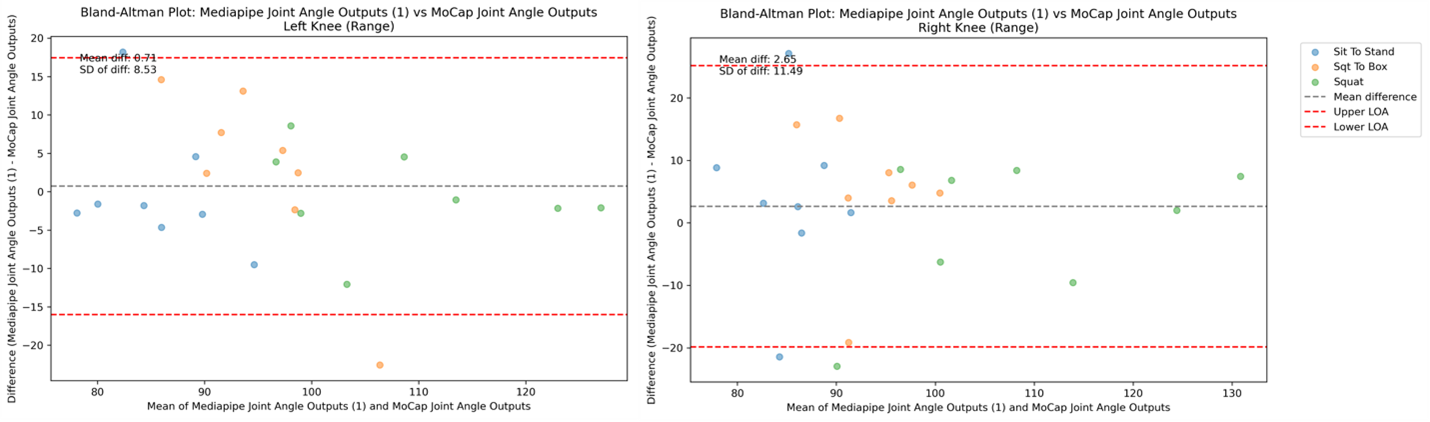}
    \caption{Bland-Altman plot illustrating the agreement between MoCap and IMU. The mean difference (bias) is shown as a solid line, while the dashed lines represent the limits of agreement (mean difference ± 1.96 standard deviations), the points represent the range of motion of both the left knee flexion (left) and right knee flexion (right) of different actions for each participant.}
    \label{fig:ba_mocap}
\end{figure}

Figures \ref{fig:ba_mediapipe} and \ref{fig:ba_mocap} show Bland Altman plots comparing two different modalities, mediapipe (\ref{fig:ba_mediapipe}) and MoCap (\ref{fig:ba_mocap}), against the gold standard of IMU-based kinematics. Both \ref{fig:ba_mediapipe} and \ref{fig:ba_mocap}, for both the left and right knee angle, show a good agreement between the two modalities and the spread between the two limits of agreement shows that there is no systematic bias. While there are some points that are beyond the limits of agreement the number of these falls within a reasonable level.

\subsection{Error Analyses}

\begin{figure}
    \centering
    \includegraphics[width=1\linewidth]{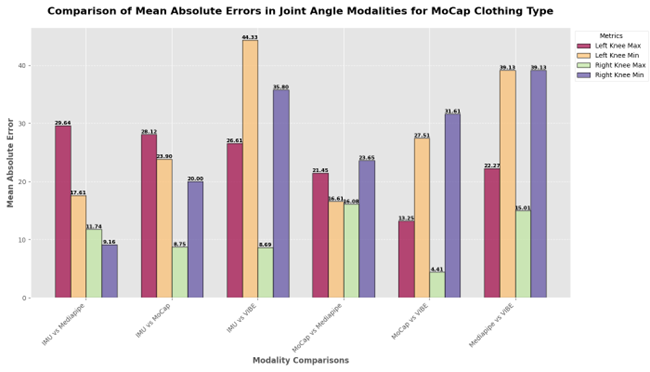}
    \caption{Comparing the mean absolute error of the minimum and maximum knee angles for the left and right knee between different modality combinations for the MoCap clothing.}
    \label{fig:mae_mocap}
\end{figure}

\begin{figure}
    \centering
    \includegraphics[width=1\linewidth]{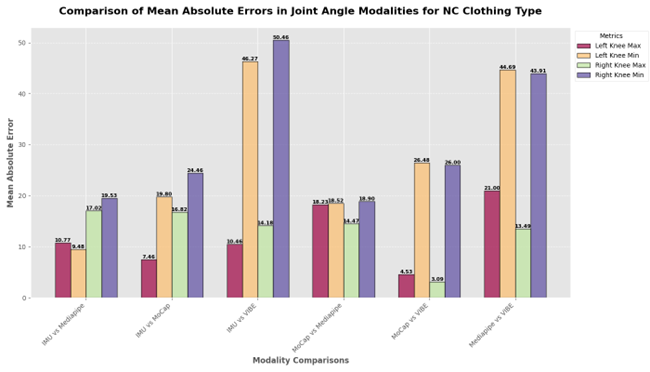}
    \caption{Comparing the mean absolute error of the minimum and maximum knee angles for the left and right knee between different modality combinations for the normal clothing.}
    \label{fig:mae_nc}
\end{figure}

Figures \ref{fig:mae_mocap} and \ref{fig:mae_nc} show the MAE comparisons of different modality pairs for the minimum and maximum knee angles for the left and right knee for both MoCap clothing (\ref{fig:mae_mocap}) and normal clothing (\ref{fig:mae_nc}). Both figures show that for mediapipe the minimum values are closer to those from the IMU than the maximum values, the difference between the MAE comparing the IMU to mediapipe and MoCap to mediapipe are very similar showing a consistent result. These results also show that the VIBE alogrithm’s kinematics are closer to MoCap than mediapipe, especially the minimum knee angle. Another observation when comparing mediapipe and MoCap, we see a smaller MAE in normal clothing than in the MoCap clothing. 

\begin{figure}
    \centering
    \includegraphics[width=1\linewidth]{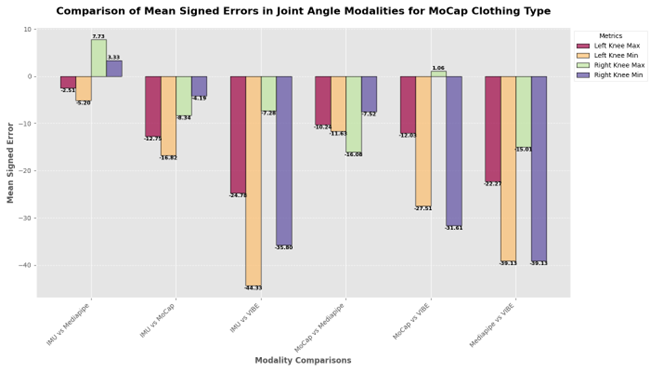}
    \caption{Comparing the mean signed error of the minimum and maximum knee angles for the left and right knee between different modality combinations for the MoCap clothing.}
    \label{fig:mse_mocap}
\end{figure}

\begin{figure}
    \centering
    \includegraphics[width=1\linewidth]{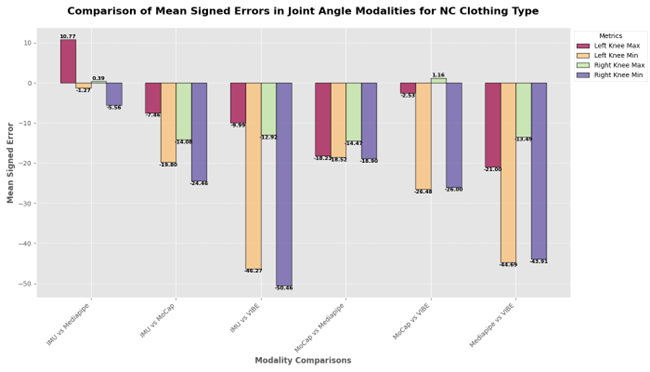}
    \caption{Comparing the mean signed error of the minimum and maximum knee angles for the left and right knee between different modality combinations for the normal clothing.}
    \label{fig:mse_nc}
\end{figure}

Figures \ref{fig:mse_mocap} and \ref{fig:mse_nc} on the other hand, show the MSE comparisons of different modality pairs of the same kinematic metrics. These results show that the mediapipe’s error is closer to the IMU than any other data collection method. However, when comparing the kinematics to MoCap rather than the IMUs you observe a smaller error in the maximum values when using VIBE and similar levels of error for the minimum values for both mediapipe and VIBE.

\subsection{Statistical Analysis}

The statistical tests used in this study aimed to show that there was no significant difference between the marker-less mediapipe joint angles and the gold standard in sports biomechanics. The results in Tables \ref{tab:mocap_imu}-\ref{tab:normal_mocap} show that for both clothing conditions and all three tested actions, there were no significant differences in the range of motion of either knee angle.

\begin{table}[h]
    \centering
    \begin{tabular}{lllrr}
        \toprule
        Action & Joint & T-Statistic & P-Value \\
        \midrule
        Sit-to-stand & Left Knee  & -0.786 & 0.462 \\
        Sit-to-stand & Right Knee & -0.729 & 0.494 \\
        Squat to Box & Left Knee  & 6.623  & 0.095 \\
        Squat to Box & Right Knee & -0.449 & 0.731 \\
        Squat        & Left Knee  & -0.794 & 0.458 \\
        Squat        & Right Knee & -0.548 & 0.604 \\
        \bottomrule
    \end{tabular}
    \caption{Paired t-test results for "MoCap" clothing comparing Mediapipe and the IMU range of motion. No kinematics with a p-value < 0.05 show that there is no significant difference between the two modalities.}
    \label{tab:mocap_imu}
\end{table}

\begin{table}[h]
    \centering
    \begin{tabular}{lllrr}
        \toprule
        Action & Joint & T-Statistic & P-Value \\
        \midrule
        Sit-to-stand & Left Knee  & 2.036 & 0.088 \\
        Sit-to-stand & Right Knee & 1.184 & 0.281 \\
        Squat to Box & Left Knee  & 0.932 & 0.404 \\
        Squat to Box & Right Knee & 0.359 & 0.738 \\
        Squat        & Left Knee  & 2.270 & 0.072 \\
        Squat        & Right Knee & 1.716 & 0.147 \\
        \bottomrule
    \end{tabular}
    \caption{Paired t-test results for "Normal" clothing comparing Mediapipe and the IMU range of motion. No kinematics with a p-value < 0.05 show that there is no significant difference between the two modalities.}
    \label{tab:normal_imu}
\end{table}

\begin{table}[h]
    \centering
    \begin{tabular}{lllrr}
        \toprule
        Action & Joint & T-Statistic & P-Value \\
        \midrule
        Sit-to-stand & Left Knee  & 0.017 & 0.987 \\
        Sit-to-stand & Right Knee & 0.399 & 0.703 \\
        Squat to Box & Left Knee  & -0.879 & 0.541 \\
        Squat to Box & Right Knee & -0.710 & 0.607 \\
        Squat        & Left Knee  & 1.071 & 0.325 \\
        Squat        & Right Knee & 0.137 & 0.895 \\
        \bottomrule
    \end{tabular}
    \caption{Paired t-test results for "MoCap" clothing comparing Mediapipe and the MoCap range of motion. No kinematics with a p-value < 0.05 show that there is no significant difference between the two modalities.}
    \label{tab:mocap_mocap}
\end{table}

\begin{table}[h]
    \centering
    \begin{tabular}{lllrr}
        \toprule
        Action & Joint & T-Statistic & P-Value \\
        \midrule
        Sit-to-stand & Left Knee  & 0.508 & 0.629 \\
        Sit-to-stand & Right Knee & 1.167 & 0.287 \\
        Squat to Box & Left Knee  & -0.131 & 0.490 \\
        Squat to Box & Right Knee & 0.953 & 0.395 \\
        Squat        & Left Knee  & -0.758 & 0.483 \\
        Squat        & Right Knee & 0.138 & 0.896 \\
        \bottomrule
    \end{tabular}
    \caption{Paired t-test results for "Normal" clothing comparing Mediapipe and the MoCap range of motion. No kinematics with a p-value < 0.05 show that there is no significant difference between the two modalities.}
    \label{tab:normal_mocap}
\end{table}

\section{Discussion}

Given the increasing demand and strain in healthcare, clinical environments require innovative and cost-effective solutions that will be less intensive than currently used methods. While a large amount of research has gone into developing new marker-less motion capture methods, these very rarely touch on the clinical applications and thus they have little validation in terms of their accuracy in extracting clinically relevant features from motion. Previous studies have identified the potential of motion data in disease prediction \cite{ibrahim2021real} and rehabilitation tracking of patients \cite{cotton2022posepipe}, both in-person and remotely \cite{li2020human}, but this needs more extensive validation in terms of the accuracy before widespread use can be achieved. 

The results presented in this study show that, when considering factors such as the ease of use, set-up time, recording time, lack of professional training required, and the initial cost, a marker-less solution is an excellent option for a clinical environment. While the error of these technologies are still high when compared to both the IMU and MoCap, which are the current state-of-the-art, each of these gold standard methods have their own inherent error. 

As clinically relevant 1D biomarkers have already been validated in a previous study and showed they can be used as a statistically significant clinical decision-making tool \cite{armstrong2024marker}, this study instead prioritised comparing the kinematics measured from different sources of motion data. The choice to use actions with well-defined and consistent peaks between participants, both the sit-to-stand and the squat to box actions, allows us to compare the angle measured between the modalities directly. This also removed the need to have a synchronised start and end of the motion, which is difficult to achieve when three modalities are not able to be directly connected to one another due to hardware limitations. 

These 1D biomarkers, minimum knee angle, maximum knee angle, and range of knee motion, while not being useful in a sports biomechanics perspective, provide useful objective data to help in the clinical decision-making process \cite{holla2012diagnostic}.  Given the purpose of this study is to determine the accuracy of these tools compared to the state-of-the-art for human motion analysis, to ultimately justify the use of these tools in a clinical setting. Further studies will be required to compare the tools when measuring more complex metrics from the field of sports biomechanics rather than these simpler 1D metrics. 
However, even though these results show the capabilities of the marker-less motion capture techniques such as Mediapipe, the gold standard to use for maximum accuracy is still MoCap and the IMUs due to both their higher recording frequencies and higher measurement accuracies. These higher frequencies allow more information to be captured from the motion, thus, allowing for the detection of smaller changes that may still have relevance to a pathology.

On the other hand, for a clinical tool, two of the key factors affecting the use in clinical environments are the speed of use and the cost. Given that the marker-less tools can be used with a simple RGB camera on either a mobile phone or portable laptop computer, tools which are already readily available to clinicians, this solution presents a cheap, fast, and portable solution that still shows a good degree accuracy when measuring clinically relevant kinematics.

Overall, the results presented in this study show that new marker-less motion capture tools such as Mediapipe have become powerful tools for the measurement of clinically relevant kinematics without the need for complex training and is able to collect data without any time-consuming set-up and calibration step. This presents a useful tool that can be deployed in either fast-paced clinical environments or to be used directly by the patient to measure kinematics remotely for monitoring of disease progression or measure improvement after intervention. However, the current gold standard tools of IMUs and marker-based MoCap are still the preferred option for a more detailed understanding of a subject's motion due to the higher overall accuracy and the higher recording frequency.

\bibliographystyle{unsrt}  
\bibliography{references} 






\end{document}